\documentclass[12pt]{article}
\usepackage{color}
\usepackage{latexsym,amsmath,amssymb,amsthm,amsfonts,graphicx,natbib,enumerate,url}
\usepackage{epsfig}
\usepackage{setspace}
\usepackage{bm}
\usepackage{psfrag,epsf}
\usepackage[outdir=./]{epstopdf}

\usepackage{adjustbox}

\usepackage[linesnumbered,ruled,vlined]{algorithm2e}
\usepackage[noend]{algpseudocode}

\SetCommentSty{mycommfont}

\SetKwInput{KwInput}{Input}                
\SetKwInput{KwOutput}{Output}              
\newcommand*\patchAmsMathEnvironmentForLineno[1]{%
      \expandafter\let\csname old#1\expandafter\endcsname\csname #1\endcsname
      \expandafter\let\csname oldend#1\expandafter\endcsname\csname end#1\endcsname
      \renewenvironment{#1}%
         {\linenomath\csname old#1\endcsname}%
         {\csname oldend#1\endcsname\endlinenomath}}%
    \newcommand*\patchBothAmsMathEnvironmentsForLineno[1]{%
      \patchAmsMathEnvironmentForLineno{#1}%
      \patchAmsMathEnvironmentForLineno{#1*}}%
    \AtBeginDocument{%
    \patchBothAmsMathEnvironmentsForLineno{equation}%
    \patchBothAmsMathEnvironmentsForLineno{align}%
    \patchBothAmsMathEnvironmentsForLineno{flalign}%
    \patchBothAmsMathEnvironmentsForLineno{alignat}%
    \patchBothAmsMathEnvironmentsForLineno{gather}%
    \patchBothAmsMathEnvironmentsForLineno{multline}%
    }
\usepackage[mathlines,displaymath]{lineno}	

\newcommand{\blind}{0}

\usepackage{ifpdf}

\begin{document}


\def\spacingset#1{\renewcommand{\baselinestretch}%
{#1}\small\normalsize} \spacingset{1}


\if0\blind
{
  \title{\bf A Bayesian Long Short-Term Memory Model for Value-at-Risk and Expected Shortfall Joint Forecasting}
  \author{Zhengkun Li, Minh-Ngoc Tran, Chao Wang, Richard Gerlach, Junbin Gao\thanks{
    This research was partially supported by the Australian Research Council (ARC) Discovery Project scheme DP200103015.}\hspace{.2cm}\\
    Discipline of Business Analytics, Business School, The University of Sydney\\}

  \maketitle
} \fi

\if1\blind
{
  \bigskip
  \bigskip
  \bigskip
  \begin{center}
    {\LARGE\bf A Bayesian Long Short-Term Memory Model for Value-at-Risk and Expected Shortfall Joint Forecasting}
\end{center}
  \medskip
} \fi

\bigskip

\begin{abstract}
Value-at-Risk (VaR) and Expected Shortfall (ES) are widely used in the financial sector
to measure the market risk and manage the extreme market movement. The recent link between
the quantile score function and the Asymmetric Laplace density has led to a flexible
likelihood-based framework for joint modelling of VaR and ES. It is of high interest in
financial applications to be able to capture the underlying joint dynamics of these two
quantities. We address this problem by developing a hybrid model that is based on
the Asymmetric Laplace quasi-likelihood and employs the Long Short-Term Memory
(LSTM) time series modelling technique from Machine Learning to capture
the underlying dynamics of VaR and ES efficiently. We refer to this model as LSTM-AL. We adopt
the adaptive Markov chain Monte Carlo (MCMC) algorithm for Bayesian inference in
the LSTM-AL model. Empirical results show that the proposed LSTM-AL model can
improve the VaR and ES forecasting accuracy over a range of well-established competing models.

\end{abstract}
\noindent%
\vfill
\newpage
\spacingset{1.45} 
\section{Introduction}\label{sec:Introduction}
Value-at-Risk (VaR) and Expected Shortfall (ES) are two risk measures that are widely used by financial institutions as tools to manage the market risk and the extreme market movement. An $\alpha$-level VaR is defined as the $\alpha$-level quantile of the underlying portfolio return distribution. The VaR has been used as the standard market risk measure for setting the regulatory capital requirement. However, VaR does not give any information in terms of the expected loss conditional on the losses beyond the VaR threshold, and thus it may fail to capture the expected extreme risk especially for assets with a fat-tail return distribution. Also, VaR is not a subadditive measure, which means the overall VaR of a well-diversified portfolio can be larger than the aggregated VaR of each individual asset. ES compliments VaR and is able to better capture the risk with fat-tail return distributions  \citep{artzner1997thinking,artzner1999coherent}. An $\alpha$-level ES is defined as the conditional expectation of exceedances beyond the corresponding $\alpha$-level VaR. Compared to VaR, ES is a more coherent measure with several attractive properties such as the subadditivity \citep{acerbi2002coherence}. Together with VaR, ES has been employed as a tail risk measure for financial regulation and recommended by the Basel Accord.

The classical statistical approaches, especially the parametric and non-parametric approaches, to estimate VaR and ES are well-established, where ES forecasts can be regarded as a by-product of VaR forecasts. Parametric models, such as the generalized autoregressive conditional heteroskedasticity (GARCH) family models \citep{engle1982autoregressive,BOLLERSLEV1986307}, make strong assumptions about the underlying return distribution and obtain the VaR estimation by modelling the volatility dynamics. Non-parametric approaches, such as the historical simulation (HS) method, forecast the VaR by the corresponding quantile of the empirical distribution, and forecast the ES by averaging the exceedances beyond VaR. 
Semi-parametric approaches, e.g. the quantile regression conditional autoregressive VaR (CAViaR) model of \cite{engle2004caviar}, directly model the underlying VaR series by adapting quantile regression and thus can avoid the distribution assumption of the return. However, quantile regression models cannot direct produce ES forecasts. \cite{taylor2008estimating} proposed a semi-parametric model named as Conditional Autoregressive Expectile (CARE) to forecast the VaR and ES jointly by incorporating the expectile.

ES is not an elicitable measure, i.e. there is no scoring or loss function that can be minimized by the true ES \citep{gneiting2011making}. This makes it difficult to develop regression-type models for estimating the ES. Nevertheless, \cite{fissler2016higher} found that ES and VaR are jointly elicitable and thereby proposed a family of scoring functions for evaluating VaR and ES forecasts jointly. \cite{taylor2019forecasting} proposed a framework using the Asymmetric Laplace (AL) density to model VaR and ES jointly, leading to a quasi-likelihood framework for performing inference and prediction, referred to here as the ES-CAViaR model.  Furthermore, \cite{patton2019dynamic} proposed a VaR and ES dynamics model based on the family of scoring functions and provided an asymptotic analysis.  

Classical statistical approaches commonly use simple models to capture the underlying latent process. For example, the stochastic volatility model uses an AR(1) process to model the latent volatility dynamics, and the GARCH model uses a simple linear combination of the previous historical volatility and squared return to model the future volatility. However, strong evidence from the literature suggests the existence of long-range and non-linear serial dependence in financial volatility (see, e.g. \citet{ding1993long,so2006empirical,kilicc2011long}). This indicates that simple processes such as AR(1) may fail to capture efficiently the complicated dynamics of the underlying volatility process.

Neural Network (NN) modeling is widely used in the Deep Learning literature as a powerful functional approximation tool with the NN structures varying depending on the data type and learning task. The Long Short-Term Memory (LSTM) structure of \cite{hochreiter1997long} is a special type of Recurrent Neural Network (RNN) which is designed for modeling the sequential data. LSTM is well-known for its ability to capture efficiently the long-term and non-linear serial dependence in time series data.
However, for financial time series modelling, \cite{makridakis2018statistical} documented that sophisticated Machine Learning models may not be able to outperform simple statistical models. The reasons can be twofold. First, the parameter estimation process will be challenging for hybrid models embedded with complex Machine Learning techniques. Second, Machine Learning techniques such as the LSTM usually use observations as both the input and output, thus how to handle unobservable latent variables such as financial volatility is also challenging. 

To the best of our knowledge, \cite{nguyen2019long} is one of a few papers that considered to model VaR by incorporating the LSTM structure into an econometric model. By combining the LSTM structure with the Stochastic Volatility (SV) model, they showed that their so-called LSTM-SV model is able to capture efficiently the volatility dynamics and reported some favourable results in terms of VaR forecast. However, this paper does not consider joint modeling of VaR and ES, and relies on the distribution assumption like other parametric models do.

Our paper proposes a semi-parametric framework, named as LSTM-AL, which incorporates the LSTM structure into the ES-CAViaR framework with a careful modification by using the latent variables for both the input and output of the LSTM structure. The proposed LSTM-AL model retains the semi-parametric property and therefore does not rely on any distribution assumption of returns. The LSTM-AL model is able to capture the possible long-term and non-linear dependence in the joint dynamics of VaR and ES. We consider a full Bayesian treatment for LSTM-AL and adopt an adaptive MCMC algorithm for Bayesian inference. Both the simulation and empirical studies well document that the LSTM-AL model can capture efficiently the non-linearity and long-term dependence exhibited in the underlying dynamics. It is demonstrated that the LSTM-AL model can provide more accurate out-of-sample forecasts than the ES-CAViaR model across a range of financial datasets.

The paper is organized as follows. Section 2 reviews the development of VaR \& ES loss functions and the associated ES-CAViaR framework. Section 3 motivates the use  of Neural Network techniques in financial econometrics and proposes the LSTM-AL model. Section 3 also presents Bayesian inference for LSTM-AL using MCMC. Section 4 highlights the property of the LSTM-AL model in capturing non-linearity and long-term dependence by designing an comprehensive simulation study. Section 5 applies the LSTM-AL model to four financial datasets and compares its performance with several well-developed models in the literature. Section 6 concludes.

\section{The Joint Scoring Function and ES-CAViaR Framework }\label{sec:joint scoring}
A scoring function refers to a loss function that can be optimized by the true prediction of some elicitable measures \citep{fissler2016higher}. VaR is an elicitable measure as the true VaR forecasts can minimize the quantile loss function, while ES is not an elicitable measure. Nevertheless, \citet{fissler2016higher} described the joint elicitability of VaR and ES and proposed a general framework of the joint scoring function. Furthermore, \citet{taylor2019forecasting} proposed the AL scoring function and the ES-CAViaR regression framework. We now describe the joint scoring functions and the ES-CAViaR framework. 

\subsection{Quantile Loss Function for VaR}\label{sec:qlf for VaR}
VaR is an elicitable measure and thus, the true VaR forecasts can minimize certain loss function, namely, quantile loss function, firstly proposed by \cite{koenker1978regression}. The quantile loss function is defined as:

\begin{equation}  \label{expression: qlf}
    \text{QL} = (r_t-\mathbf{VaR}_t)(\alpha-\mathbf{I}_{(r_t<\mathbf{VaR}_t)}),
\end{equation}
where $r_t$ is the return value at time $t$ and $\mathbf{VaR_t}$ is the $\alpha$ level quantile at time $t$. Averaging or summing the quantile loss across a sample gives the measure for evaluating the VaR quantile forecast performance. This quantile loss function has been widely employed in quantile regression for VaR forecasting (e.g., the CAViaR framework of \cite{engle2004caviar}).

\subsection{Joint Loss Function for VaR and ES}\label{sec:AL for VaR and ES}
ES is not elicitable \citep{gneiting2011making}, therefore no loss function can be used to evaluate ES forecasting performance. However, \cite{fissler2016higher} described the joint elicitability of VaR and ES, that is, the true VaR and ES forecasts can optimize a scoring function jointly. As the result, they specified the general form of VaR and ES joint scoring function, which is defined as:
\begin{equation}
    \begin{split}
    \mathcal{S}(\mathbf{VaR}_t,\mathbf{ES}_{t}, r_t, \alpha) &= (\mathbf{I}_{(r_t\leq\mathbf{VaR}_t)}-\alpha)G(\mathbf{VaR}_{t}) - \mathbf{I}(r_t\leq\mathbf{VaR}_t)G(r_t)\\ &+ \zeta'(\mathbf{ES}_{t})\Big(\mathbf{ES}_t-\mathbf{VaR}_t+\mathbf{I}_{(r_t\leq\mathbf{VaR}_{t})} \times (\mathbf{VaR}_t - r_t)/\alpha\Big) - \zeta(\mathbf{ES}_{t}) + a(r_t),\label{Joint score}
\end{split}
\end{equation}
where $G$ and $\zeta$ are functions that must satisfy certain conditions, including that $G$ is increasing and $\zeta$ is increasing and convex. Based on this expression, \cite{taylor2019forecasting} proposed the AL log score function to test VaR \& ES jointly by taking into account both the magnitude of violation and the violation (hit) ratio, which is defined as:

\begin{equation}
    \text{AL}(\mathbf{VaR}_t,\mathbf{ES}_{t}, r_t, \alpha) =  -\ln\Big(\frac{\alpha-1}{\mathbf{ES}_t}\Big) - \frac{(r_t-\mathbf{VaR}_t)(\alpha-\mathbf{I}_{(r_t\leq\mathbf{VaR}_t)})}{\alpha\mathbf{ES}_t}.\label{AL}
\end{equation}
Averaging or summing the AL score across a sample will return the measure for joint evaluating VaR and ES forecast performance. The joint scoring function allows using optimization frameworks to estimate VaR and ES. An example is Taylor's ES-CAViaR regression framework that will be described next.

\subsection{ES-CAViaR Framework}\label{sec:ES-CAViaR}
\cite{taylor2019forecasting} described several models in order to capture various effects exhibited in the joint dynamics of VaR and ES. For the VaR component, two CAViaR specifications \citep{engle2004caviar} were proposed. The first is Symmetric Absolute Value (SAV):
\begin{equation}\label{SAV}
    \mathbf{VaR}_{t} = \beta_0 + \beta_1\mathbf{VaR}_{t-1} + \beta_2|r_{t-1}|,
\end{equation}
and the second is Asymmetric Slope (AS):
\begin{equation}\label{AS}
    \mathbf{VaR}_{t} = \beta_0 + \beta_1\mathbf{VaR}_{t-1} + \beta_2(r_{t-1})^{+} + \beta_3(r_{t-1})^{-},
\end{equation}
where $(x)^{+} = \text{max}(x,0)$, $(x)^{-} = -\text{min}(x,0)$ and the $\mathbf{\beta}$ are the parameters. The SAV framework uses a GARCH-type transition to model the VaR dynamics, while the AS takes into account the well-known leverage effect in finance, i.e. a negative innovation tends to have a larger impact on volatility than a positive innovation. For ES, \cite{taylor2019forecasting} proposed two frameworks. The first is Exponential (EXP):
\begin{equation}\label{multiple}
    \mathbf{ES}_{t} = \Big(1+\exp(\gamma_0)\Big)\mathbf{VaR}_{t},
\end{equation}
which models the dynamics of ES by multiplying $\mathbf{VaR}$ by a factor greater than 1 to avoid the crossing between ES and VaR. The EXP framework might be too restrictive as it does not allow much flexibility for the dynamics of ES. To address this, \citeauthor{taylor2019forecasting} proposed the Mean Exceedance (EXC):
\begin{align}
    \label{me1}&\mathbf{ES}_{t} = \mathbf{VaR}_{t} - x_t\\\label{me2}
    &x_t = \left\{
    \begin{array}{cc}\gamma_0+\gamma_1(\mathbf{VaR}_{t-1}-r_{t-1})+\gamma_2x_{t-1} & \text{if}~ r_{t-1}\leq\mathbf{VaR}_{t-1}\\
    x_{t-1} & \text{otherwise}
    \end{array}
\right.,
\end{align}
which models the non-negative difference between VaR and ES by a autoregressive process.

In our empirical study in Section \ref{sec:Empirical}, we use two combinations of the above frameworks, SAV-EXP and AS-EXC, to form two benchmark models to compare with our new approach.
\section{The LSTM-AL model}\label{sec:LSTM-AL}
The ES-CAViaR approach can model the joint dynamics of VaR and ES and is shown to outperform a range of competing models \citep{taylor2019forecasting}. An extended framework that incorporates realized measures is also proposed recently \citep{gerlach2020semi}. However, the simple linear dynamics frameworks in ES-CAViaR may fail to capture the non-linearity and long-term serial dependence exhibited in financial volatility. This section describes our flexibly hybrid approach that incorporates the LSTM structure into ES-CAViaR to capture effectively the complicated joint dynamics of VaR and ES.
\subsection{RNN and LSTM structures}
The statistical time series models commonly use the linear regression-type approach to model the target variable. For example, the well-known AR process uses the autoregression linear function to model the future value of the target with the lagged observations from the previous time steps. This linear regression-type approach is simple and works well in many cases, but might fail to simulate efficiently complicated dynamics that exhibits non-linear and long-term serial dependence.  

For time series data, the recurrent neural network (RNN) modeling technique in the Machine Learning literature is well known for its capacity to describe complicated underlying structures in sequential data \citep{Goodfellow2016}.
Let $\{{x}_{t}, t = 1,2,\ldots\}$ be the input sequential data and 
$\{{y}_{t}, t = 1,2,\ldots\}$ be the output sequential data.
The goal is to model the conditional distribution of $y_t$, given input $x_t$ and the information up to time $t-1$. 
The most simple RNN model, known as Elman's model \citep{Elman}, is
\begin{align}
    &h_t = \sigma(\mu x_t+\omega h_{t-1}+b),\;\;h_0=0,\\
    &\eta_t = \beta_0+\beta_1h_t,\\
    &y_t|\eta_t \sim p(y_t|\eta_t),\;\; t=1,2,...
\end{align}
where $\mu,\omega,b,\beta_0$ and $\beta_1$ are model parameters. The hidden unit $h_t$ is updated recurrently and stores the memory from the previous time steps, thus allows the RNN structure to be able to capture the serial dependence in the underlying time series. $\sigma(\cdot)$ is the non-linear activation function (e.g., sigmoid or tanh) which allows the RNN structure to capture the non-linearity exhibited in the underlying sequential data. The conditional distribution $p(y_t|\eta_t)$ needs to be specified depending on the learning task. Notice that  both the input ${x}$ and output ${y}$ can be real-valued vectors. For the purpose of this paper we only consider ${x}$ and ${y}$ as scalar. 

However, there are two significant drawbacks in the simple RNN structure \citep{bengio1994learning}. First, it is difficult for the simple RNN to learn the long-term dependence. Second, the simple RNN is subject to the well-known gradient exploding or vanishing issues, as the gradient may either explode or approach to zero. To address these issues, the Long Short-Term Memory (LSTM) structure was proposed by \cite{hochreiter1997long} which extends the basic RNN structure with three extra hidden units, namely, input gate $g_t^i$, output gate $g_t^o$ and forget gate $g_t^f$, to mitigate the gradient problem and control the information flow to capture the long-term dependencies. The LSTM structure can be expressed as LSTM($x_t,h_{t-1}$) with the following equations:
\begin{align}
    &\text{Forget Gate} &g^f_t = \sigma(\mu_fx_t+\omega_fh_{t-1}+b_f)\\
    &\text{Input Gate} &g^i_t = \sigma(\mu_ix_t+\omega_ih_{t-1}+b_i)\\
    &\text{Data Input} &x^d_t = \sigma(\mu_dx_t+\omega_dh_{t-1}+b_d)\\
    &\text{Output Gate} &g^o_t = \sigma(\mu_ox_t+\omega_oh_{t-1}+b_o)\\
    &\text{Cell State} &C_t=g^f_t\odot C_{t-1}+g^i_t\odot x^d_t\\
    &\text{Date Output} &h_t = g^o_t\odot\text{tanh}(C_t)
\end{align}
where $\sigma(\cdot)$ is the sigmoid activation function and $\text{tanh}(\cdot)$ is the tanh activation function. $\odot$ refers to the element-wise multiplication. The parameter set of a general LSTM structure is $\{\mu_f,\omega_f,b_f,\mu_i,\omega_i,b_i,\mu_d,\omega_d,b_d,\mu_o,\omega_o,b_o\}$. With the proposal of the cell state $C_t$, the LSTM architecture can mitigate the gradient problem. Also, the introduction of the three new gates, $g_t^f, g_t^i$ and $g_t^o$, guarantees the LSTM can keep the important information output, $h_{t-1}$, from the previous time steps and also forgets the useless content. This structure is extremely useful as the information stored in the cell state can flow continuously through the time. As the result, the LSTM model can capture the long-term dependence as well as the non-linearity. The power of LSTM structure has been well documented in many Deep Learning applications including language translation, video data processing, etc. For more details related to the RNN and LSTM architectures, we refer the interested readers to \cite{lipton2015critical} and \cite{Goodfellow2016}.

\subsection{The LSTM-AL Model}
\begin{figure}[h]
\begin{center}
\includegraphics[height=9cm,width=16cm]{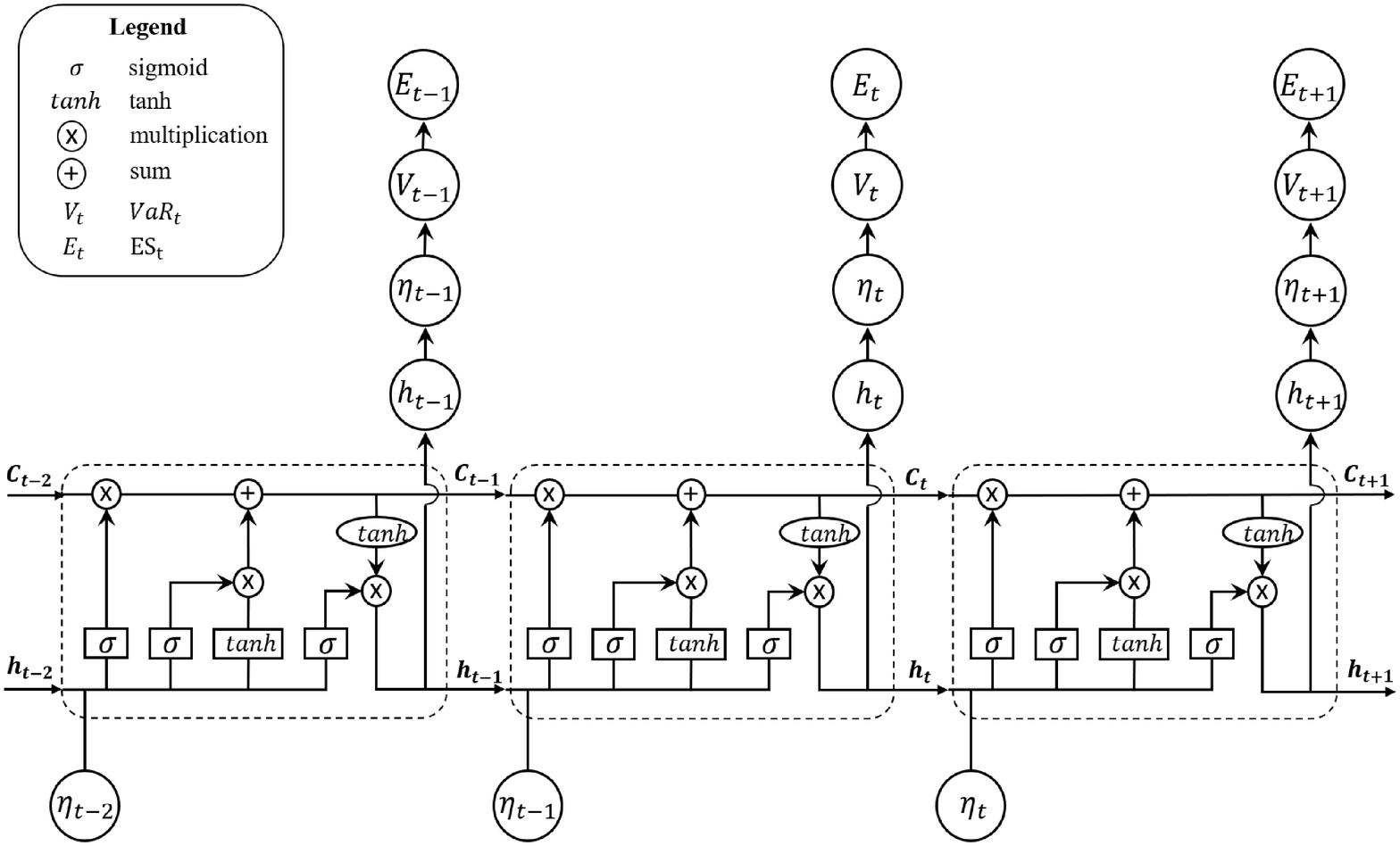}
\caption{Graphical representation of the LSTM-AL model in Expression. (\ref{LSTM-AL:VaR})-(\ref{LSTM-AL:LSTM})}
\label{fig: LSTM-AL Structure}

\end{center}
\end{figure}

In order to flexibly model the joint dynamics of VaR and ES, this section proposes an innovative LSTM-AL model by combining the ES-CAViaR framework with the LSTM structure. The LSTM-AL model is defined as:
\begin{align}
    &p(r_t|\mathbf{VaR}_t,\mathbf{ES}_{t}) = \frac{\alpha-1}{\mathbf{ES}_t}\exp\Big(\frac{(r_t-\mathbf{VaR}_t)(\alpha-\mathbf{I}_{(r_t\leq\mathbf{VaR}_t)})}{\alpha\mathbf{ES}_t}\Big)\\
    &\mathbf{VaR}_t = \eta_t+\beta_0|r_{t-1}|+\beta_1\mathbf{VaR}_{t-1}\label{LSTM-AL:VaR}\\
    &\mathbf{ES}_t = \big(1+\exp(\gamma_0+\gamma_1 h_t)\big)\mathbf{VaR}_t\label{LSTM-AL:ES}\\
    &\eta_t = \alpha_0 + \alpha_1h_t\label{LSTM-AL:eta}\\
    &h_t = \text{LSTM}(\eta_{t-1}, h_{t-1}),\label{LSTM-AL:LSTM}
\end{align}
where the parameter set $\theta$ includes $\{\beta_0, \beta_1, \gamma_0, \gamma_1, \alpha_0, \alpha_1\}$ and the 12 parameters in the LSTM structure. We employ the SAV formulation from \cite{taylor2019forecasting} with a small modification, where we model the drift term $\eta_t$ with the LSTM architecture to capture the possible long-term and non-linear serial dependence of the VaR dynamics. Also, as discussed in Section \ref{sec:ES-CAViaR}, the EXP framework might be too restrictive as it does not allow much flexibility for the dynamics of ES, we add the output of the LSTM structure, $h_t$, into the ES formulation of expression (\ref{LSTM-AL:ES}), to allow for more flexibility in the ES dynamics. 
Figure \ref{fig: LSTM-AL Structure} presents the graphical representation of the LSTM-AL model, inspired by \cite{nguyen2019long}.
It's worth noting that the SAV-EXP model is a special case of our LSTM-AL model when $\alpha_1=\gamma_1=0$. 

It is straightforward to extend the LSTM-AL model in multiple ways. First, by incorporating the other VaR formulations (e.g., the AS formulation of expression (\ref{AS})), the model can capture different types of effects exhibited by the underlying time series. Also, the realized measures can be incorporated to further improve the model's ability in capturing latent volatility efficiently (e.g., see \cite{gerlach2016forecasting,wang2019bayesian}). We leave these potential extensions for future research. 
\subsection{Bayesian Inference}
This section describes Bayesian inference for the LSTM-AL model. The favourability of Bayesian inference in financial time series applications has been well documented in the literature \citep{gerlach2011bayesian,wang2019semi}. Despite using the sophisticated LSTM structure, the LSTM-AL model only has 18 parameters and we use adaptive MCMC to sample from their posterior distribution \citep{roberts2009examples, haario2001adaptive}.
Other Bayesian estimation methods such as Sequential Monte Carlo or Variational Bayes can be employed too.

For the proposed LSTM-AL model, the likelihood function is tractable (AL likelihood). Algorithm \ref{MCMC algo} describes the adaptive MCMC which can automatically turn the covariance matrix of the proposal distribution to enhance the convergence of the MCMC chain and also target the theoretically optimal acceptance rate of 23.4\% \citep{roberts2009examples}. 

\begin{algorithm}[h]
\DontPrintSemicolon
  
  \KwInput{Initial parameter set $\theta_0$} 
  \KwOutput{Posterior samples of the LSTM-AL parameters}
  \For{$n\leq2d$}
  {Sample $\theta^*$ from the proposal density $\mathcal{N}(\theta_n, \sigma_{ini}\mathbf{I}_d)$
  
  Compute the acceptance probability $\alpha(\theta^*,\theta_n) = \text{min}\big\{1, \frac{p(r_{i:j}|\theta^*)p(\theta^*)}{p(r_{i:j}|\theta_n)p(\theta_n)}\big\}$, and sample $u\sim U(0,1)$
  
  \If{$\alpha(\theta^*,\theta_n)<u$}{\text{Accept} $\theta^*$ and update $\theta_{n+1} = \theta^*$}
  \Else{$\theta_{n+1}=\theta_n$}
  }
  
  \For{$n>2d$}
    {Sample $\theta^*$ from $(1-\beta)\mathcal{N}(\theta_n,\sigma_{opt}c_n)+\beta\mathcal{N}(\theta_n,\sigma_{ini}\mathbf{I}_d)$
  
  Compute the acceptance probability $\alpha(\theta^*,\theta_n) = \text{min}\big\{1, \frac{p(r_{i:j}|\theta^*)p(\theta^*)}{p(r_{i:j}|\theta_n)p(\theta_n)}\big\}$
  
  \If{$\alpha(\theta^*,\theta_n)<u$}{\text{Accept} $\theta^*$ and update:
  
    $\theta_{n+1} = \theta^*$
  
    $m_{n+1} = \frac{n}{n+1}m_n+\frac{1}{n+1}\theta_{n}$
  
    $c_{n+1} = \frac{n-1}{n}c_n + m_n^2+\frac{1}{n}\theta_{n+1}^2 - \frac{n-1}{n}m_{n+1}^2$}
   
   \Else{$\theta_{n+1}=\theta_n$}
   }
   
\caption{Adaptive MCMC Algorithm}
\label{MCMC algo}
\end{algorithm}
In Algorithm \ref{MCMC algo}, $m_n$ and $c_n$ refer to the empirical estimate of the mean and covariance matrix, $\sigma_{ini} = \frac{0.1^2}{d}$ and $\sigma_{opt} = \frac{2.38^2}{d}$ are the initial and optimal scales of the covariance matrix with $d$ the size of $\theta$ \citep{roberts2009examples}. We used $\beta = 0.05$ in our implementation. Table \ref{LSTM-AL prior information} lists the prior distributions used in the LSTM-AL model. We follow \cite{gerlach2020semi} to employ the flat prior distribution for parameters $\beta_0,\beta_1,\gamma_0$ and $\gamma_1$ from the VaR \& ES transition equations. On the other hand, we follow \cite{nguyen2019long} and use a normal prior for the 12 LSTM parameters.

\begin{table}[htbp]
  \centering
    \begin{tabular}{cccccccc}
    \hline
    \hline
    \multicolumn{8}{c}{LSTM-AL} \\
    \hline
    Parameter & $\beta_0$ & $\beta_1$ & $\gamma_0$ & $\gamma_1$ & $\alpha_0$ & $\alpha_1$ & LSTM \\
    Prior & Flat  & Flat  & Flat  & Flat  & $\mathcal{N}(0,0.1)$ & IG(2.5,0.25) & $\mathcal{N}(0,0.1)$ \\
    \hline
    \hline
    \end{tabular}%
      \caption{Prior distributions for the Bayesian LSTM-AL model.}

  \label{LSTM-AL prior information}%
\end{table}%

\section{Simulation Study}\label{sec:Simulation}
An intensive simulation study is conducted to compare both the in-sample and out-of-sample properties of the proposed Bayesian LSTM-AL model, in comparison with a range of well-developed competing models.

\begin{figure}[h]
\begin{center}
\includegraphics[height=6cm, width=16cm]{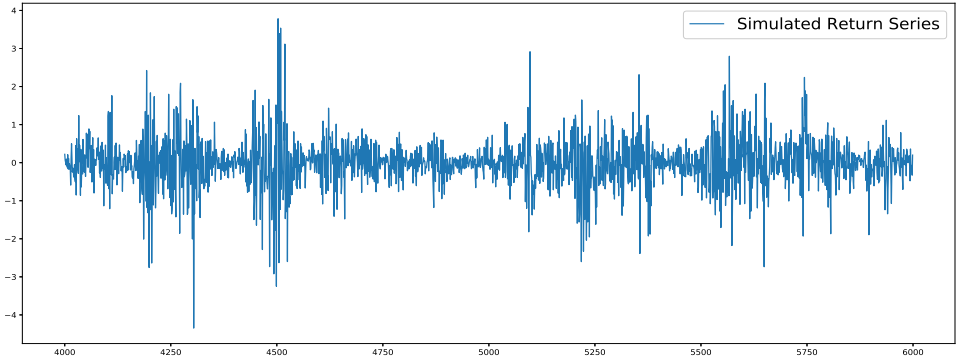}
\caption{Example of the simulated nonlinear return series}
\label{fig: simulated data series plot}

\end{center}
\end{figure}

Following \citet{nguyen2019long}, ten series of returns are simulated from the following non-linear stochastic volatility model:
\begin{align}
    &v_t =  0.1 + 0.96v_{t-1}-0.8\frac{v_{t-1}^2}{1+v_{t-1}^2}+\frac{1}{1+\exp(-v_{t-1})}+\sqrt{0.1}\epsilon^v_{t},~t=2,\cdots,T,\\
    &r_t = \exp(\frac{1}{2}v_t)\epsilon_t^r,~ t = 1,\ldots,T\\
    &\epsilon^v_t, \epsilon^r_t \sim\mathcal{N}(0,1),
\end{align}
where $\{v_t\}$ denotes the log volatility process and $\{r_t\}$ denotes the return series. The initial volatility value $v_1$ is randomly sampled from $\mathcal{N}(0,1)$. Each time we generated 6000 observations and kept the last 2000 observations as the final simulated data. Figure \ref{fig: simulated data series plot} provides an example of the simulated series where the well-known financial characteristics such as volatility clustering are observable.

\begin{table}[h]
  \centering
    \begin{tabular}{ccccccc}
    \hline
    \hline
          & mean  & std   & min  & max  & skew  & kurt \\
    \hline
    $\gamma_0$    & -1.0231 & 2.2246 & -4.9982 &  4.9958 & 0.2763 & -0.6523 \\
    $\gamma_1$     & -0.1880 & 2.7067 & -5.0000 & 4.9959 &  0.0746& -1.0920 \\
    $\beta_0$     & -2.5857 & 1.2849 & -4.9993 & 0.4784 & 0.0247 & -0.9882 \\
    $\beta_1$     & 0.4561& 0.2528 & -0.1815& 0.9978 &-0.0122 & -0.9347 \\
    $\alpha_0$     & -0.2826 & 0.2083 &  -1.1673 & 0.3828 &  -0.5226 &  0.4221 \\
    $\alpha_1$     & 0.1909 & 0.1932 & 0.0187 & 2.0192 &  3.4996 & 17.0949 \\
    $\mu_f$     & 0.0041 & 0.3217 &  -1.1944 & 1.0803 & 0.0118 & -0.0870 \\
    $\omega_f$     & -0.0145 & 0.3180 & -1.2674 & 1.1214 &  -0.1205 & -0.0211 \\
    $b_f$     & -0.0023 & 0.3207 & -1.2821 & 1.1148 & -0.0459 &  0.0362 \\
    $\mu_i$     &0.0058 &  0.3206 & -1.1791 & 1.3240 &  -0.0504 &  -0.0024 \\
    $\omega_i$    & -0.0006 & 0.3263 & -1.1724 &  1.3421 & -0.0145 & 0.0651 \\
    $b_i$    & -0.0079 & 0.3227 & -1.3108 & 1.0774 & -0.0237 & 0.0700 \\
    $\nu_d$    & -0.0022& 0.3249 & -1.1286 & 1.1640 & 0.0513 & -0.0743 \\
    $\omega_d$    & -0.0243 & 0.3242 & -1.2550 &  1.1701 &  0.0128 &  0.1524 \\
    $b_d$    & -0.0091 & 0.3105 & -1.1681 &  1.1372& 0.0665 &  -0.0113\\
    $\nu_o$    & 0.00001 &0.3186 &  -1.1174 & 1.3202 &  0.0071 &  -0.0169\\
    $\omega_o$    & -0.0050 & 0.3155 & -1.0760 & 1.0270 & 0.0070 & -0.1457 \\
    $b_o$    &  0.0040 & 0.3293 & -1.2405 &1.2192 &-0.0443 & -0.0517 \\
    \hline
    \hline
    \end{tabular}%
      \caption{Posterior descriptive statistics of the Bayesian LSTM-AL parameters}

  \label{tab:LSTM_para_S1}%
\end{table}%

With each set of the simulated data, we used the first 1000 observations for Bayesian model estimation (in-sample analysis) and the last 1000 observations for model evaluation (out-of-sample analysis). We compare the performance of the LSTM-AL model with two ES-CAViaR family models, SAV-EXP and AS-EXC. Each model was estimated by the adaptive MCMC Algorithm \ref{MCMC algo} which was run for 50,000 iterations with the first 15,000 discarded as the burn-ins to guarantee the Markov Chain converge. For the SAV-EXP and AS-EXC models, the prior distributions were set to be flat as in \cite{gerlach2020semi}.

Table \ref{tab:LSTM_para_S1} lists the estimated parameters of the LSTM-AL model from one of the ten simulated datasets. The estimated means for $\alpha_1$  and $\gamma_1$ significantly differ from zero, which indicates that the LSTM-AL model is able to capture the non-linearity exhibited in the underlying volatility dynamics of the simulated data.

\begin{table}[htbp]
  \centering

    \begin{tabular}{ccc}
\hline
\hline
    \multicolumn{3}{c}{1\% VaR \& ES} \\
\hline
          & Quantile Loss & AL Score \\
    LSTM-AL & \textbf{0.0346} & \textbf{2.1862} \\
    SAV-EXP & 0.0411 & 2.4035 \\
    AS-EXC & 0.0450 & 2.5002 \\
\hline
\hline
    \end{tabular}%
      \caption{Averaged loss function values across the ten simulated series at 1\% quantile level}
      \label{tab: simulation loss function}%
\end{table}%

For the out-of-sample study, we rolling the one-step-ahead forecasts with the parameter estimated by the MCMC with a window size of 1,000 for each model, across the ten simulated datasets. To access the model performance in VaR \& ES forecasting, we employ the quantile loss function for VaR and the AL score function for VaR \& ES jointly as both criteria considered the magnitude of violation and the violation rate.

\begin{figure}[h]
\begin{center}
\includegraphics[height=8cm, width=16cm]{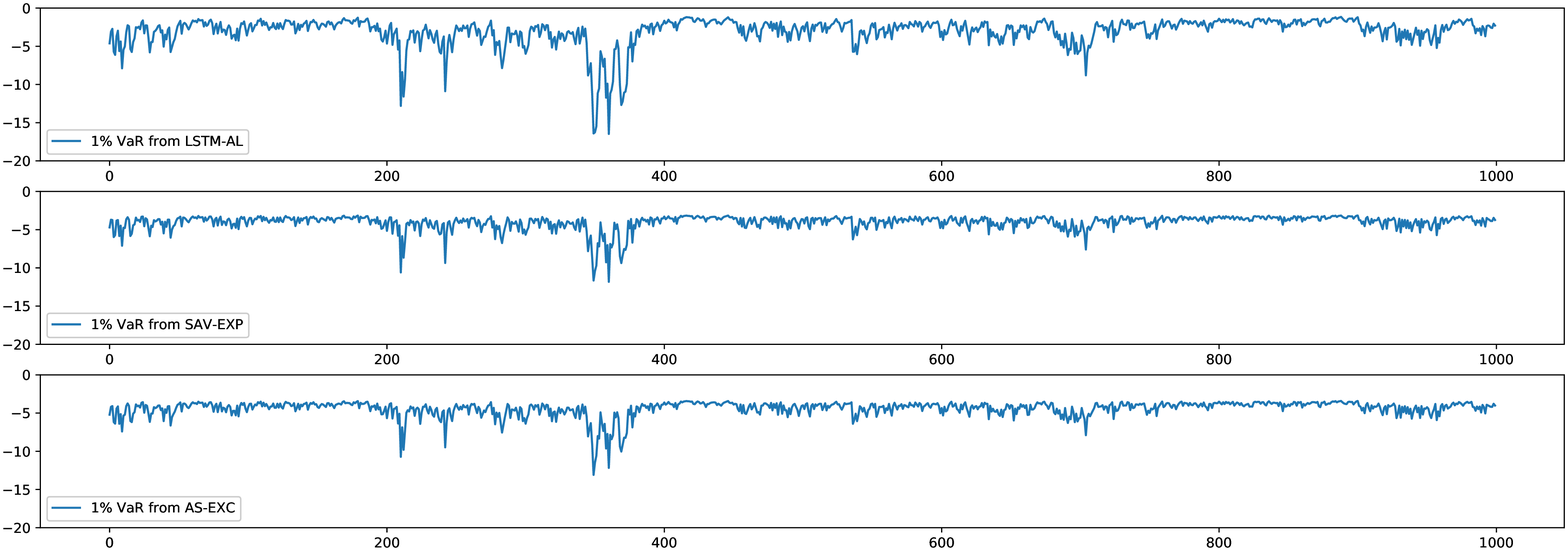}
\caption{1\% VaR dynamics for the nonlinear return series in Figure 2}
\label{fig: VaR simulation}
\end{center}
\end{figure}

\begin{figure}[h]
\begin{center}
\includegraphics[height=8cm, width=16cm]{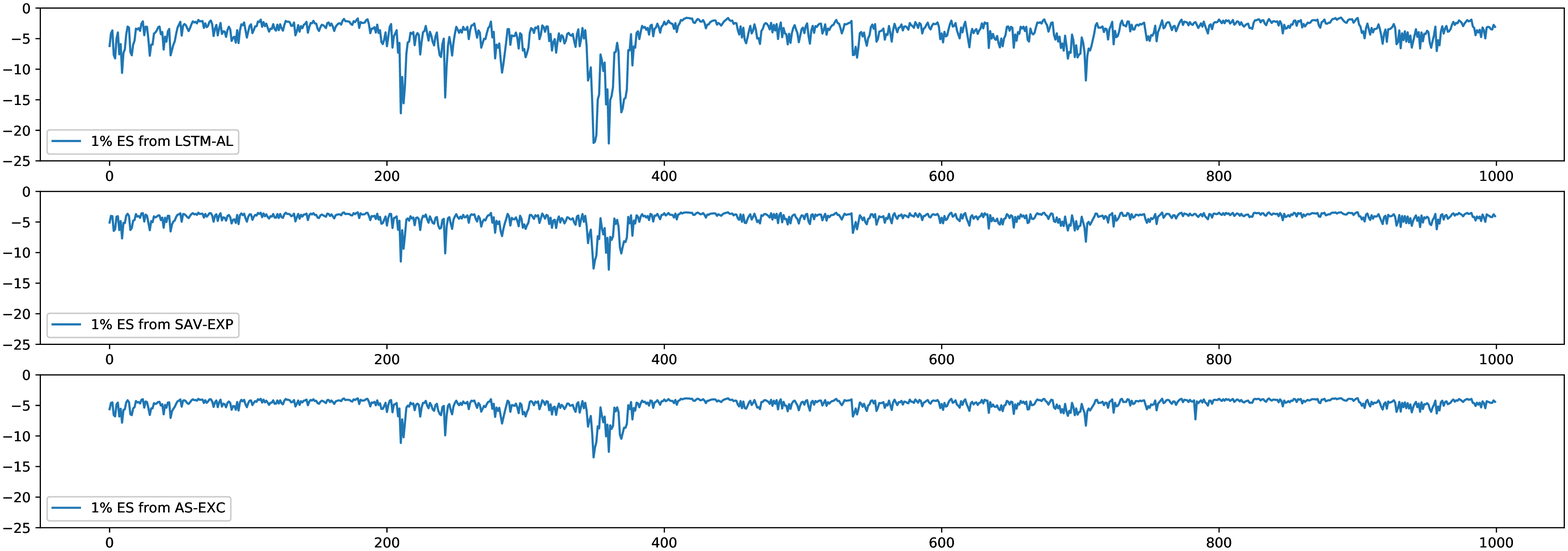}
\caption{1\% ES dynamics for the nonlinear return series in Figure 2}
\label{fig: ES simulation}
\end{center}
\end{figure}

Table \ref{tab: simulation loss function} summarizes the performance measure averaged over the ten datasets which shows that the LSTM-AL model outperforms the SAV-EXP and AS-EXC models. We note that the proposed LSTM-AL model outperforms two benchmark models in all the ten simulated datasets in terms of the quantile loss and AL score. To visualize the joint forecasts of VaR \& ES, we plot the out-of-sample forecasting series for the first simulated data in Figure \ref{fig: VaR simulation} \& \ref{fig: ES simulation}.

The simulation study highlights the favourability of the proposed LSTM-AL model. The empirical evidence from next section strongly supports this conclusion further.
\section{Empirical Study}\label{sec:Empirical}
This section presents the empirical study where we applied the proposed LSTM-AL model into four financial time series to test its ability in capturing VaR and ES dynamics.
\begin{figure}[h]

\begin{center}
\includegraphics[height=10cm, width=16cm]{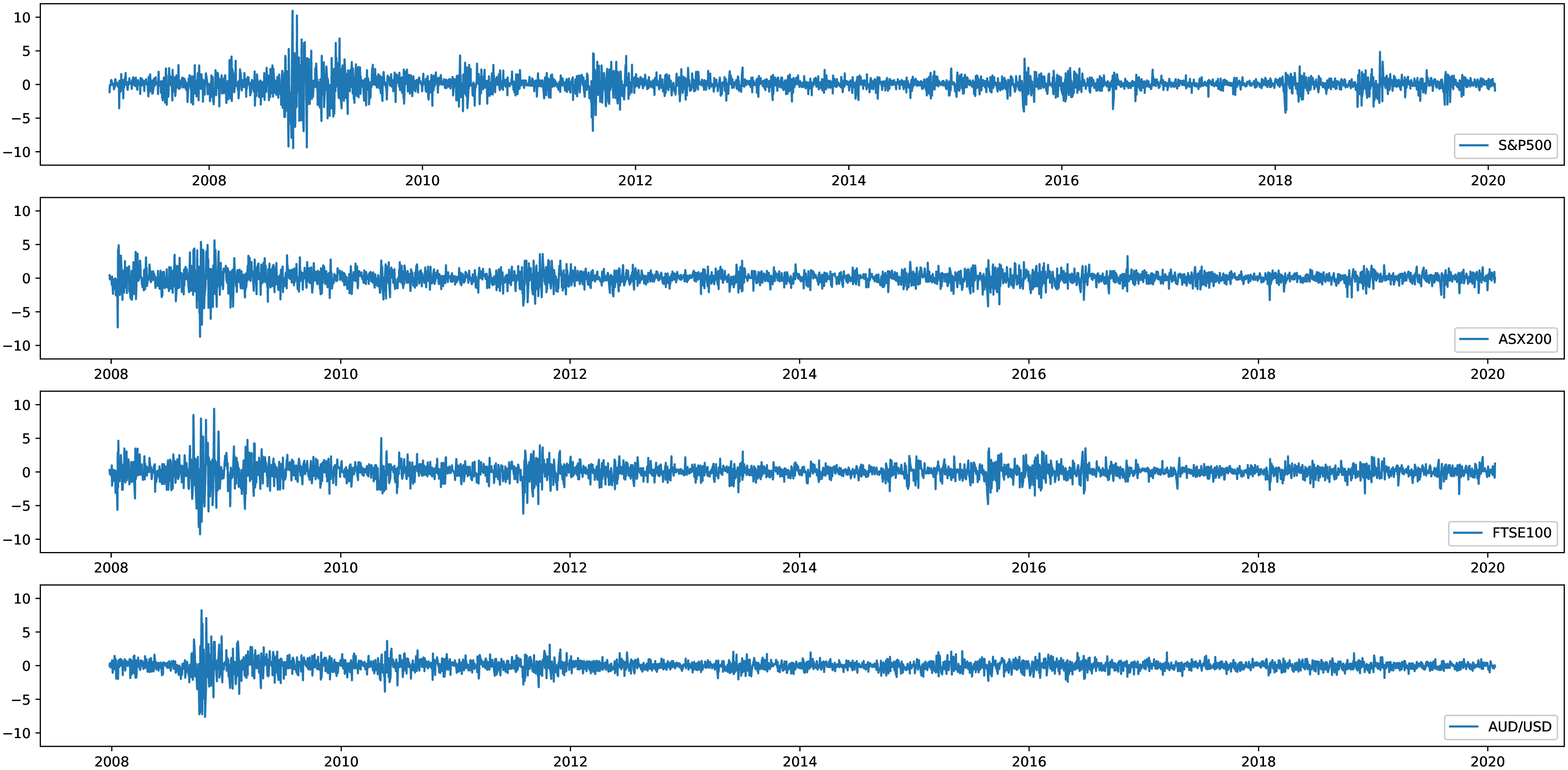}
\caption{Time Series Plots of S\&P500, ASX200, FTSE100 and AUDUSD}
\label{data series plot}
\end{center}

\end{figure}

\subsection{Data Description}
Daily closing prices data from four markets: S\&P500 (US), ASX200 (Australia), FTSE100 (UK) and AUD/USD (exchange rate), are collected from Yahoo Finance to evaluate the proposed model and a range of competitors. With the available closing prices, we obtain the percentage log return series:
\begin{equation}
    r_t = \Big(\ln(P_t) - \ln(P_{t-1})\Big) \times 100,
\end{equation}
where $P_t$ denotes the closing price at time $t$. Some descriptions of these datasets are listed in Table \ref{data information} where we use the first 2000 observations for the in-sample estimation and analysis while the rest are used for the out-of-sample analysis and a rolling window with fixed in-sample size is employed for estimation to produce each one step-ahead forecast in the out-of-sample period. The time series plots of these four markets are presented in Figure \ref{data series plot} where the financial characteristics such as the volatility clustering are observable. Similar to the simulation study in Section \ref{sec:Simulation}, we applied Taylor's SAV-EXP and AS-EXC models to compare with the performance of the LSTM-AL model.
Each model was estimated by the adaptive MCMC Algorithm \ref{MCMC algo} which was run for 50,000 iterations with the first 15,000 iterations discarded as burn-ins.

\begin{table}[htbp]
  \centering
    \begin{tabular}{cccccc}
    \hline
    \hline
          & Start & End   & Size  & In-sample Size & Out-of-sample Size \\
    \hline
    S\&P500 & 25/01/2007 & 24/01/2020 & 3272  & 2000  & 1272 \\
    ASX200 & 27/12/2007 & 28/08/2019 & 3105  & 2000  & 1105 \\
    FTSE100 & 27/12/2017 & 24/01/2020 & 3051 & 2000 & 1051 \\
    AUD/USD & 25/12/2007 & 24/01/2020 & 3154  & 2000  & 1154 \\
    \hline
    \hline
    \end{tabular}%
      \caption{S\&P500, ASX200, FTSE100 and AUD/USD datasets descriptions.}

  \label{data information}%
\end{table}%

We conduct the rolling window forecasting study over the four markets in terms of both 2.5\% and 1\% confidence levels according to the proposal of Basel Accord which moves the quantitative risk metrics system from VaR to ES and pays more attention on 97.5\% confidence level. 
Table \ref{tab: emp loss function results1} \& \ref{tab: emp loss function results25} lists values for both the quantile loss function and AL log score function.

\begin{figure}[h]

\begin{center}
\includegraphics[height=8cm, width=16cm]{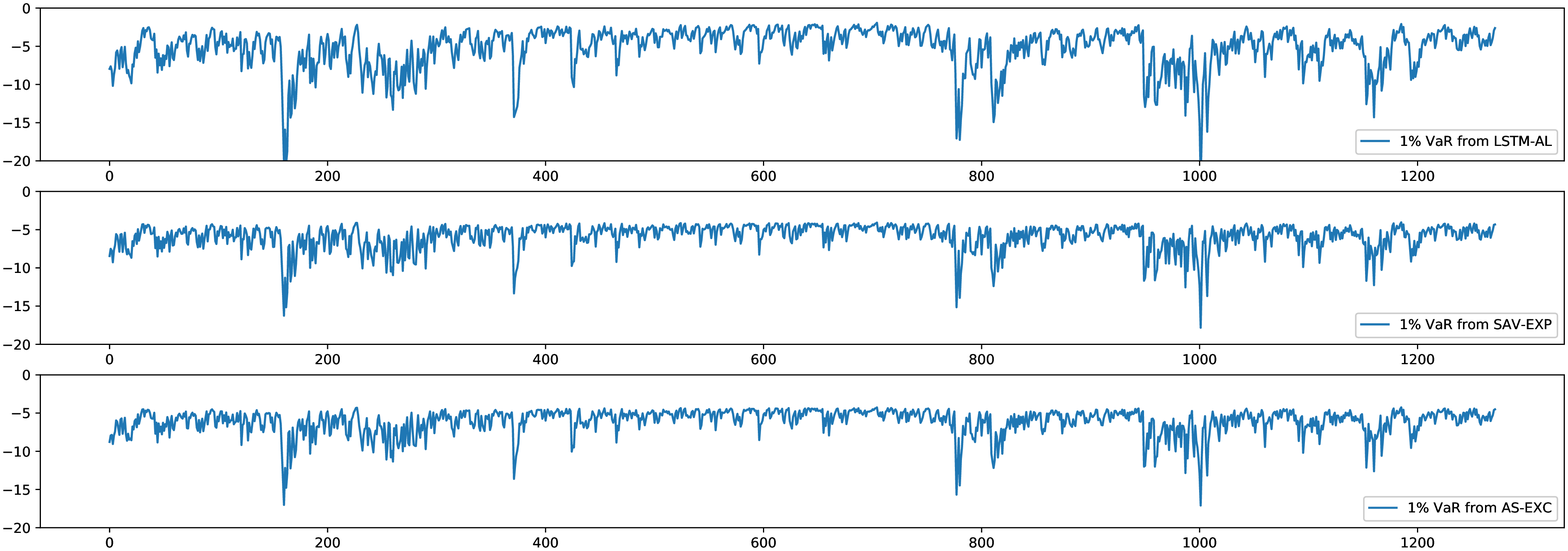}
\end{center}
\caption{1\% VaR dynamics forecasting, evidence from S\&P500 market.}
\label{plot: VaR dynamics}
\end{figure}

\begin{figure}[h]

\begin{center}
\includegraphics[height=8cm, width=16cm]{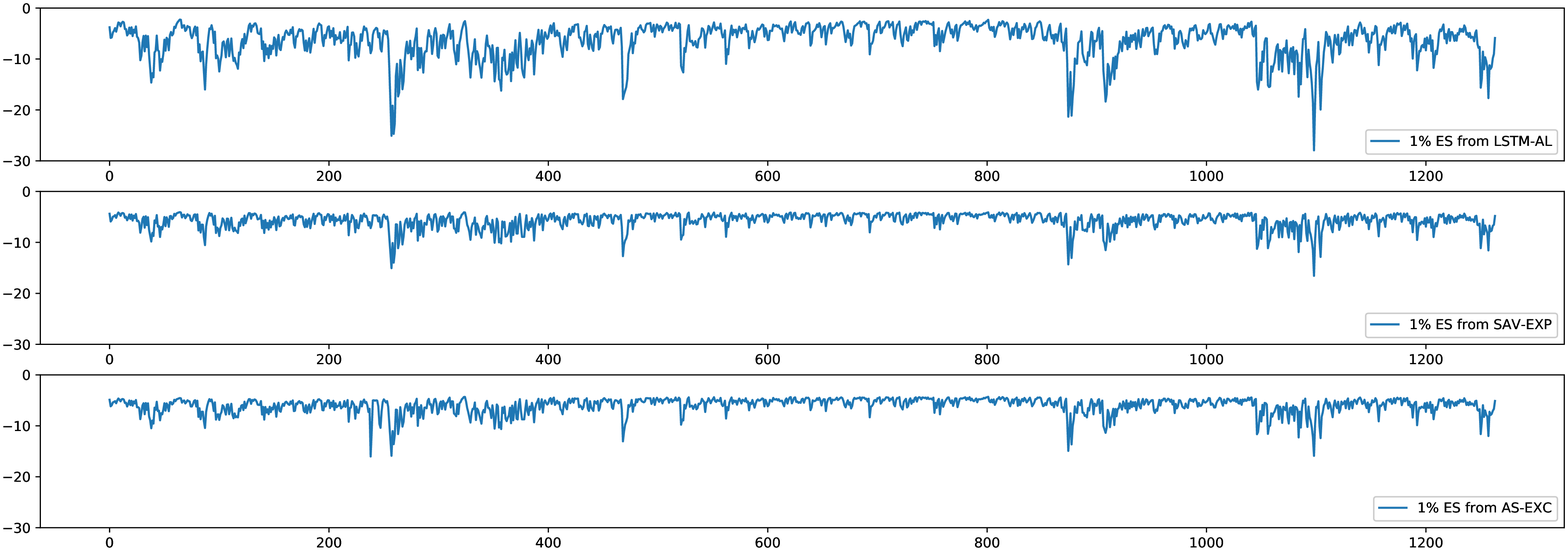}
\end{center}
\caption{1\% ES dynamics forecasting, evidence from S\&P500 market.}
\label{plot: ES dynamics}
\end{figure}

The overall performance of the proposed LSTM-AL model is promising.
At both the 2.5\% and 1\% level forecasting, the LSTM-AL model consistently outperforms the five benchmark models across all the three stock index markets. This indicates that the proposed LSTM-AL model can capture well the non-linear and long-term serial dependence properties exhibited by the selected financial stock markets. For the only foreign exchange market, GJRGARCH-t and EGARCH-t provide very competitive results in terms of the quantile loss function and AL score, while the LSTM-AL model can still provide the second best forecasting result at both 2.5\% and 1\% levels. To visualize the forecasting dynamics, we presents the 99\% confidence level VaR \& ES forecasting dynamics for the S\&P500 market in Figure \ref{plot: VaR dynamics} and \ref{plot: ES dynamics}.

\begin{table}[htbp]
  \centering
    \begin{adjustbox}{width=\columnwidth,center}
    \begin{tabular}{ccccccccc}
    \hline    
    \hline    
    \multicolumn{9}{c}{1\% VaR \& ES} \\
    \hline    
          & \multicolumn{4}{c}{Quantile Loss} & \multicolumn{4}{c}{AL Score} \\
    \hline    
          & \multicolumn{1}{c}{S\&P500} & \multicolumn{1}{c}{ASX200} &
          \multicolumn{1}{c}{FTSE100} &
          \multicolumn{1}{c}{AUD/USD} & \multicolumn{1}{c}{S\&P500} & \multicolumn{1}{c}{ASX200} &
          \multicolumn{1}{c}{FTSE100} & \multicolumn{1}{c}{AUD/USD} \\
    \hline
 
        GARCH-t &0.1015 & 0.1009&0.0992 &0.0903 & 2.5126&2.4214 & 2.5077&2.4311 \\
    GJRGARCH-t &0.1009 & 0.0958&0.0942 &0.0891 &2.5113 & 2.4107&2.4596 & 2.4011\\
    EGARCH-t &0.1011 & 0.0921&0.0957 & \textbf{0.0831}& 2.5358&2.3912 & 2.4722&2.3722 \\
    LSTM-AL & \textbf{0.0974} & \textbf{0.0910} & \textbf{0.0935} & 0.0836 & \textbf{2.5051} & \textbf{2.3851} &\textbf{2.4284} & \textbf{2.3083} \\
    SAV-EXP & 0.1021 & 0.1016& 0.1013 & 0.0887 & 2.5639 & 2.4807 &2.5214 &2.5838 \\
    AS-EXC & 0.1025& 0.1047 &0.0977 &0.0914 & 2.5230 & 2.4922& 2.5017 & 2.5983 \\
   \hline
    \hline    
    
    \end{tabular}%
         \end{adjustbox}

      \caption{Out-of-sample loss function values across the four   markets.}

      \label{tab: emp loss function results1}%
\end{table}%

\begin{table}[htbp]
  \centering
    \begin{adjustbox}{width=\columnwidth,center}
    \begin{tabular}{ccccccccc}
    \hline    
    \hline    

\multicolumn{9}{c}{2.5\% VaR \& ES} \\
    \hline    
          & \multicolumn{4}{c}{Quantile Loss} & \multicolumn{4}{c}{AL Score} \\
    \hline    
          & \multicolumn{1}{c}{S\&P500} & \multicolumn{1}{c}{ASX200} &
          \multicolumn{1}{c}{FTSE100} &
          \multicolumn{1}{c}{AUD/USD} & \multicolumn{1}{c}{S\&P500} & \multicolumn{1}{c}{ASX200} &
          \multicolumn{1}{c}{FTSE100} &
          \multicolumn{1}{c}{AUD/USD} \\
    
    GARCH-t & 0.1006& 0.0921& 0.0952& 0.0902& 2.5426&2.4223 & 2.4624&2.4832 \\
    GJRGARCH-t & 0.0982 & 0.0903&0.0897 & \textbf{0.0757}&2.5126 &2.4017 &2.4177 & 2.4228\\
    EGARCH-t & 0.0994& 0.0872&0.0924 &0.0832 & 2.5865& 2.3962& 2.4203&\textbf{2.3017} \\

    LSTM-AL & \textbf{0.0898} & \textbf{0.0858} &\textbf{0.0882}& 0.0801 & \textbf{2.4732} & \textbf{2.3950}&\textbf{2.4114}& 2.3021 \\
    SAV-EXP & 0.0919 & 0.0950 &0.0977& 0.0925 & 2.6680 & 2.4713 &2.4822& 2.5196 \\
    AS-EXC & 0.0976 & 0.0981 &0.0933& 0.0993 & 2.6915 & 2.5004 &2.4422& 2.6668 \\
    
    \hline    
    \hline    
    \end{tabular}%
         \end{adjustbox}

      \caption{Out-of-sample loss function values across the four   markets.}

      \label{tab: emp loss function results25}%
\end{table}%

\section{Conclusion}\label{sec:conclusion}
This paper proposes a new semi-parametric model called LSTM-AL for modeling the VaR \& ES joint dynamics. The Bayesian inference for the LSTM-AL model was performed using adaptive MCMC. With the embedding of LSTM structure, the proposed LSTM-AL model generated favourable results in both simulation study and empirical study, in terms of the VaR quantile loss function and VaR-ES joint score function, especially in the financial stock markets where the long-term dependence and non-linearity properties are present. The more accurate forecasting results can allow financial institutions to allocate their capital asset efficiently for avoiding extreme market movement as instructed by the Basel Capital Accord.

It is possible to extend the proposed hybrid framework in several ways using advances from both the deep learning and volatility modeling literature.
For example, other RNN structures rather than LSTM can be used, and different specifications from the ES-CAViaR framework can be employed to capture different characteristics of the financial markets.  
Also, incorporating the realized measures into the input of the LSTM can add further information into its cell states and thus improve the model performance.

\newpage
\bibliographystyle{apalike}
\bibliography{LSTM-AL_paper}

\end{document}